\documentclass[sn-mathphys,Numbered]{sn-jnl}

\usepackage{graphicx}%
\usepackage{multirow}%
\usepackage{amsmath,amssymb,amsfonts}%
\usepackage{amsthm}%
\usepackage{mathrsfs}%
\usepackage[title]{appendix}%
\usepackage{xcolor}%
\usepackage{textcomp}%
\usepackage{manyfoot}%
\usepackage{booktabs}%
\usepackage{algorithm}%
\usepackage{algorithmicx}%
\usepackage{algpseudocode}%
\usepackage{listings}%

\theoremstyle{thmstyleone}%
\theoremstyle{thmstyletwo}%

\theoremstyle{thmstylethree}%

\begin{document}

\title[Article Title]{Reservoir Computing with Generalized Readout based on Generalized Synchronization}


\author[1]{\fnm{Akane} \sur{Ookubo}}

\author*[1,2,3]{\fnm{Masanobu} \sur{Inubushi}}\email{inubushi@rs.tus.ac.jp}
\equalcont{These authors contributed equally to this work.}


\affil*[1]{\orgdiv{Department of Applied Mathematics}, \orgname{Tokyo University of Science}, \orgaddress{\city{Shinjuku}, \postcode{162-8601}, \state{Tokyo}, \country{Japan}}}

\affil[2]{\orgdiv{Graduate School of Engineering Science}, \orgname{Osaka University}, \orgaddress{\city{Toyonaka}, \postcode{560-8531}, \state{Osaka}, \country{Japan}}}

\affil[3]{\orgdiv{Department of Applied Mathematics and Theoretical Physics}, \orgname{University of Cambridge}, \postcode{CB3 0WA}, \state{Cambridge}, \country{United Kingdom}}



\abstract{
Reservoir computing is a machine learning framework that exploits nonlinear dynamics, exhibiting significant computational capabilities.
One of the defining characteristics of reservoir computing is its low cost and straightforward training algorithm, i.e. only the readout, given by a {\it linear} combination of reservoir variables, is trained.
Inspired by recent mathematical studies based on dynamical system theory, in particular generalized synchronization, we propose a novel reservoir computing framework with {\it generalized} readout, including a nonlinear combination of reservoir variables.
The first crucial advantage of using the generalized readout is its mathematical basis for improving information processing capabilities. Secondly, it is still within a linear learning framework, which preserves the original strength of reservoir computing.
In summary, the generalized readout is naturally derived from mathematical theory and allows the extraction of useful basis functions from reservoir dynamics without sacrificing simplicity.
In a numerical study, we find that introducing the generalized readout leads to a significant improvement in accuracy and an unexpected enhancement in robustness for the short- and long-term prediction of Lorenz chaos, with a particular focus on how to harness low-dimensional reservoir dynamics.
A novel way and its advantages for physical implementations of reservoir computing with generalized readout are briefly discussed.
}

\keywords{Reservoir Computing, Generalized Synchronization, Echo State Property}



\maketitle
\section{Introduction}\label{sec1}

Reservoir Computing (RC) is a machine learning framework that exploits dynamical systems~\cite{jaeger2001echo, Mass, nakajima2021reservoir}. RC has remarkable computational capabilities. For example, using random dynamical networks, RC, called echo state networks (ESNs), efficiently predicts a chaotic time series~\cite{jaeger2004harnessing}.
Adding closed-loop self-feedback makes an RC system autonomous and capable of replicating chaotic attractors, which is utilized to estimate Lyapunov exponents~\cite{pathak2017using}.
Furthermore, recent studies have shown that such `automated' RC systems can reproduce true dynamical properties more accurately than those computed from limited training data, and can extrapolate true dynamical structures such as bifurcation outside training data~\cite{kobayashi2021dynamical,kim2021teaching,Kokubu2022learning}.
Another branch of research, physical RC, harnesses various physical dynamics and has demonstrated its high performance in information processing
\cite{sunada2019photonic,appeltant2011information,wang2023echo,takano2018compact,tanaka2019recent}.

Why does RC work so well with untrained random networks and physical systems? This is a central open problem in RC research, and more broadly in machine learning or neuroscience. Partial answers to the above problem have been provided using dynamical systems theory~\cite{grigoryeva2021chaos,Kokubu2022learning,inubushi2021characteristics,inubushi2017reservoir}.
In particular, Grigoryeva, Hart, and Ortega~\cite{grigoryeva2021chaos} rigorously proved the existence of the continuously differentiable synchronization map under certain conditions and, based on this, explicitly showed what the RC learns when predicting chaotic dynamics.
In other words, they gave a formal expression of a map, which is ${\bf h}$ as explained later in the equation (\ref{def_h}), that RC approximates for prediction.
Hara and Kokubu~\cite{Kokubu2022learning} uncovered a key mathematical structure for learning with RC, i.e. a smooth conjugacy between target and reservoir dynamics, based on observations from the numerical study of the logistic map.

Inspired by these seminal works~\cite{grigoryeva2021chaos,Kokubu2022learning}, we propose a novel method of RC with a {\it generalized} readout.
Based on generalized synchronization, the Taylor expansion of the map ${\bf h}$ implies that the computational capabilities of RC with generalized readout are superior to those of conventional RC with linear readout.
Numerical studies on the Lorenz chaos prediction as a benchmark problem strongly support this; i.e., for both short- and long-term prediction, we reveal the significant computational capabilities of RC with generalized readout, mainly what we call {\it quadratic}-form RC, compared to conventional RC.
Moreover, regarding long-term prediction, we find that the automated RC system with generalized readout acquires notable robustness, in contrast to the lack of robustness of the conventional RC.

\section{Formulation}\label{sec2}

\subsection{Conventional RC}\label{sec2}

Here we briefly sketch the method of conventional RC.
Let us consider the target input ${\bf x}_{t}\in \mathbb{R}^{K}$, the target output ${\bf y}_{t}\in \mathbb{R}^{L}$ vectors ($t \in \mathbb{Z}$), and their sequence $\{ {\bf x}_{t}, {\bf y}_{t} \}$.
The goal is to construct a machine that, given an input ${\bf x}_{t}$,
produces an output $\hat{\bf y}_{t}$ that approximates the target output ${\bf y}_{t}$, i.e. $\hat{\bf y}_{t} \simeq {\bf y}_{t}$, by using the training data $\{ {\bf x}_{t}, {\bf y}_{t} \}$.


The machine consists of {\it reservoir} variables, ${\bf r}_{t} \in \mathbb{R}^{N}$, whose dynamics are determined by a map
${\bf F}: \mathbb{R}^{N} \times \mathbb{R}^{K} \to \mathbb{R}^{N}$ 
and the input $ {\bf x}_{t}$ as follows
\begin{align}
{\bf r}_{t} = {\bf F}({\bf r}_{t-1}, {\bf x}_{t}). \label{abstrc}
\end{align}
The output $\hat{\bf y}_{t}$ is determined by the {\it readout weight} matrix $W \in M_{L\times (N+1)}$ as $\hat{\bf y}_{t} = W {\bf r}_{t}$, i.e. the {\it linear} combination of the reservoir variables,
\begin{align}
(\hat{\bf y}_{t})_{i} := \sum_{j=0}^{N} W_{ij} ({\bf r}_{t})_{j} =W_{i0} + \sum_{j=1}^{N}W_{ij} ({\bf r}_{t})_{j} =  (W {\bf r}_{t})_{i}~~~(i=1, \cdots,L) \label{conv_out},
\end{align}
where we set $({\bf r}_{t})_{0} \equiv 1$ used for the approximation of the constant bias of the target output ${\bf y}_{t}$.
The readout weight matrix $W^{\ast}$ is determined such that $\hat{\bf y}_{t} \simeq {\bf y}_{t}$;
\begin{align}
W^{\ast} &= {\arg \min}_{W} \langle \| {\bf y}_{t} - \hat{\bf y}_{t} \|^{2} \rangle_{T}= {\arg \min}_{W} \langle \| {\bf y}_{t} - W {\bf r}_{t} \|^{2} \rangle_{T},
\end{align}
as usual the method of least squares,
where $\langle \cdot \rangle_{T}$ denotes the long-term average.
For simplicity, the reguralization term is omitted in the formulation, but it is used, as explained in the following numerical study.

\begin{figure}[t]%
\centering
\includegraphics[width=0.9\textwidth]{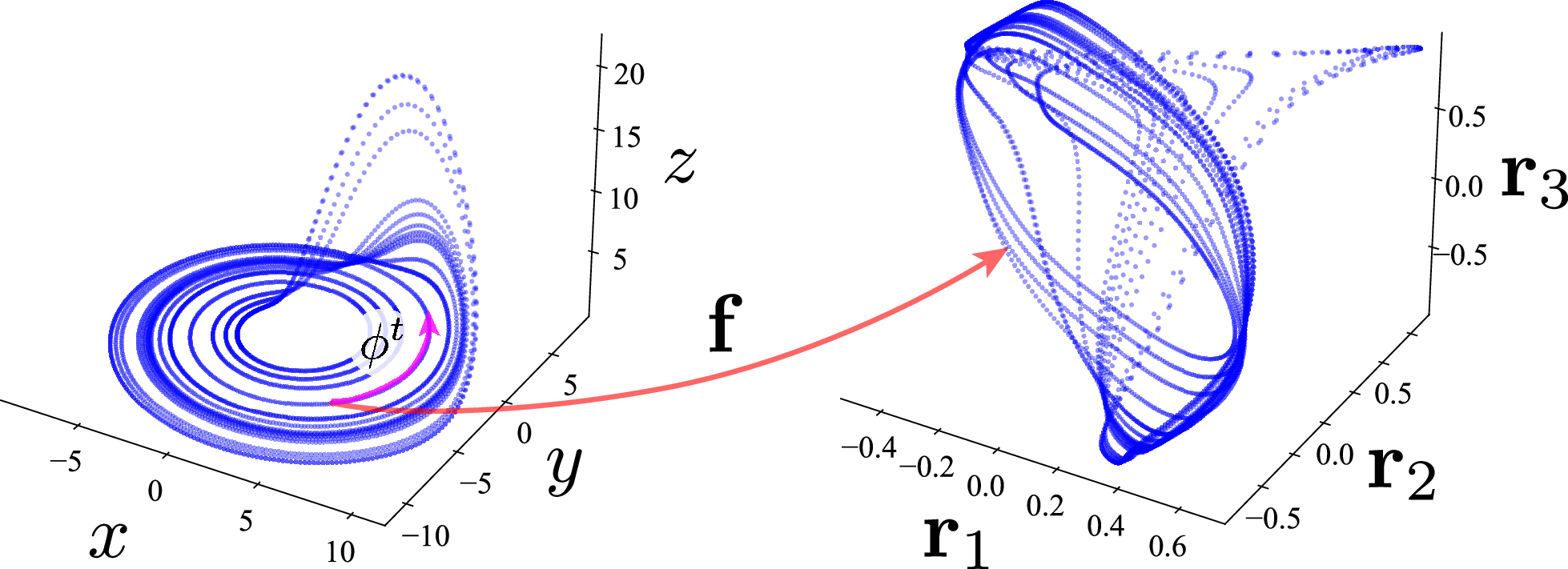}
\caption{{\bf An illustration of the target and reservoir dynamics in phase space}. As an example of the target dynamical system, the R\"{o}ssler attractor is shown in the left panel. The right panel shows the projection of the reservoir dynamics driven by the R\"{o}ssler dynamics
onto the subspace spanned by the first three variables, i.e. ${\bf r}_{1},~{\bf r}_{2},~{\bf r}_{3}$.
The red arrow depicts the schematic of the generalized synchronization map ${\bf f}$.}
\label{fig1}
\end{figure}

\subsection{Synchronizations}

Common-Signal-Induced Synchronization (CSIS),
or equivalently the Echo State Property (ESP) in the context of RC, is a key property required for reservoir dynamics determined by the map ${\bf F}$.
For a given (common) input signal $ \{ {\bf x}_{t} \}_{t \ge 1}$ and arbitrary initial reservoir states ${\bf r}_{0}, \tilde{\bf r}_{0} \in \mathbb{R}^{N}~({\bf r}_{0} \neq \tilde{\bf r}_{0})$, we say that CSIS occurs if 
the reservoir states converge to a unique state that depends only on the sequence of the input signal, i.e.
\begin{align}
\lim_{t \to +\infty} \| {\bf r}_{t} - \tilde{\bf r}_{t} \| = 0,
\end{align}
where these dynamics are determined by
\begin{align}
{\bf r}_{t} = {\bf F}({\bf r}_{t-1}, {\bf x}_{t}),~~~ \tilde{\bf r}_{t} = {\bf F}(\tilde{\bf r}_{t-1}, {\bf x}_{t}).
\end{align}
The occurrence of CSIS can be characterized by the conditional Lyapunov exponent~\cite{inubushi2021characteristics,inubushi2017reservoir}.

In this paper, we only study the input signal $ \{ {\bf x}_{t} \}_{t \ge 1}$ is generated by another dynamical system,
referred to as the {\it target} dynamical system determined by the map $\boldsymbol{\phi}: \mathbb{R}^{K} \to \mathbb{R}^{K}$,
\begin{align}
{\bf x}_{t} ={\boldsymbol{\phi}}^{t} ({\bf x}_{0}), \label{targetdynamics}
\end{align}
where ${\bf x}_{0} \in \mathbb{R}^{K}$ denotes the initial point of the target dynamics
and $\boldsymbol{\phi}^{t}=\boldsymbol{\phi} \circ \cdots \circ \boldsymbol{\phi}$ is the $t$ times composition of $\boldsymbol{\phi}$.
It is more general to formulate the observation function of the target dynamics,
$\omega$, as in Grigoryeva, Hart, and Ortega (2021)~\cite{grigoryeva2021chaos};
however, we do not consider it for simplicity.

Note that, if CSIS occurs, the asymptotic states of the reservoir dynamics $\{ {\bf r}_{t} \}$ is uniquely determined by the target dynamics $\{ {\bf x}_{t} \}$ after the transient period.
This correspondence is referred to as {\it generalized synchronization}, and denoted by
\begin{align}
{\bf r}_{t} = {\bf f}( {\bf x}_{t}),
\end{align}
where ${\bf f}: \mathbb{R}^{K} \to \mathbb{R}^{N}$ is the generalized synchronization map.
See Fig.~1 a for an illustration of ${\bf f}$.
Grigoryeva, Hart, and Ortega (2021) proved the existence and the differentiability of the map ${\bf f}$ under certain conditions~\cite{grigoryeva2021chaos}.

\subsection{Generalized RC}

Let us consider the inverse of the map ${\bf f}$ exists, and then, ${\bf x}_{t} = {\bf f}^{-1}({\bf r}_{t})$.
In that case, for instance, $\tau$-ahead prediction task of the target dynamics can be expressed by
\begin{align}
{\bf x}_{t+\tau} =\boldsymbol{\phi}^{\tau}({\bf x}_{t})= \boldsymbol{\phi}^{\tau}( {\bf f}^{-1}({\bf r}_{t}) ) =: {\bf h} ({\bf r}_{t}) \label{def_h}
\end{align}
as a function of the reservoir state ${\bf r}_{t}$.
Therefore, predicting $\tau$-ahead target dynamics with (conventional) RC is mathematically equivalent to the functional approximation of the map ${\bf h}$ using the {\it linear} combination of the reservoir variables, i.e.,
\begin{align}
{\bf h} ({\bf r}_{t}) \simeq W {\bf r}_{t}.
\end{align}
However, the map ${\bf h}$ is not linear in general. The Taylor expansion of ${\bf h}$,
\begin{align}
{\bf h}_{i}({\bf r}) = {\bf h}_{i}({\bf 0}) + \sum_{j=1}^{N} \frac{\partial {\bf h}_{i}({\bf 0})}{\partial {\bf r}_{j}}
{\bf r}_{j} + \frac{1}{2}
\sum_{j=1}^{N}\sum_{k=1}^{N} \frac{\partial^{2} {\bf h}_{i}({\bf 0})}{\partial {\bf r}_{j} \partial {\bf r}_{k}} {\bf r}_{j} {\bf r}_{k} + o(\| {\bf r} \|^{2}), \label{Taylor}
\end{align}
suggests an interpretation that 
the readout weights matrix of the conventional RC, $W$, approximate as
\begin{align}
W_{i0} \simeq {\bf h}_{i}({\bf 0}) ~~\text{and}~~ W_{ij} \simeq \frac{\partial {\bf h}_{i}({\bf 0})}{\partial {\bf r}_{j}}.
\end{align}

In this paper, we propose to utilize the {\it nonlinear} combination of the reservoir variables for the approximation of higher order terms in the Taylor expansion.
In other words, our method, referred to as {\it generalized} RC, approximates the general term in the Taylor expansion beyond the linear term, the conventional RC.
Taking into account up to second order, 
we include the quadratic form ${\bf r}_{t}^{T} W^{\mathcal{Q}} {\bf r}_{t}$
into the output as
\begin{align}
\hat{\bf y}_{t} = W {\bf r}_{t} + {\bf r}_{t}^{T} W^{\mathcal{Q}} {\bf r}_{t}
\end{align}
so that
the readout weight tensor, $W^{\mathcal{Q}} \in M_{L\times N \times N}$, approximates the Hessian term,
\begin{align}
W_{ijk}^{\mathcal{Q}} \simeq \frac{1}{2} \frac{\partial^{2} {\bf h}_{i}({\bf 0})}{\partial {\bf r}_{j} \partial {\bf r}_{k}}.
\end{align}
As the conventional RC, 
the readout weights $W^{\ast}$ and $W^{\mathcal{Q} \ast}$ are
determined such that
\begin{align}
(W^{\ast}, W^{\mathcal{Q} \ast})&= {\arg \min}_{W,W^{\mathcal{Q}}} \langle \| {\bf y}_{t} - \hat{\bf y}_{t} \|^{2} \rangle_{T} \\
&= {\arg \min}_{W,W^{\mathcal{Q}}} \langle \| {\bf y}_{t} - W {\bf r}_{t} - {\bf r}_{t}^{T} W^{\mathcal{Q}} {\bf r}_{t} \|^{2} \rangle_{T},
\end{align}
which we refer to as {\it quadratic-form} RC (QRC).

Note that learning in our method is {\it linear} with respect to the weights, $W_{ij}$ and $W^{\mathcal{Q}}_{ijk}$, which result in again least squares, and therefore, retains the simplicity of the conventional RC, i.e., the low computational cost and guaranteed optimality.
Furthermore, the output of our method is {\it nonlinear} with respect to the reservoir variables as $({\bf r}_{t})_{i} ({\bf r}_{t})_{j}$, which leads to a greater variety of approximations to the functional dependence of ${\bf r}$ on ${\bf h}({\bf r})$.
Moreover, it is natural to expect that including higher terms, beyond QRC, will give a better approximation.

\section{Numerical study of QRC and beyond}

We numerically show that the QRC is superior to the conventional RC for the Lorenz chaos prediction task, and that the closed-loop long-term prediction using the QRC provides better performance with unexpected robustness. 

Here, we use the Echo State Network (ESN) as a reservoir, i.e., the map ${\bf F}$ is given by ${\bf F}({\bf r}, {\bf x}) = \tanh ( A{\bf r} + B {\bf x})$,
where the component-wise application of $\tanh$ is employed as the activation function,
$A \in M_{N \times N}$, and $B \in M_{N \times K}$ are the random matrices.
The components of the random matrices $A$ and $B$ are sampled 
independently and identically
from a uniform random distribution over the interval 
$[-\sigma_{A}, \sigma_{A}]$ and $[-\sigma_{B}, \sigma_{B}]$, respectively,
and
we use the Ridge regression at the training phase with the regularization parameter $\beta$,
with the values given below.

The target dynamical system is determined by the Lorenz equations;
$\dot{x} = 10(y-x),~~\dot{y}= 28x - y - xz,~~\dot{z} = xy - 8/3 z,$
where its time-$\tau$ map gives $\boldsymbol{\phi}$ of the target dynamics (\ref{targetdynamics}) with some initial point ${\bf x}_{0}=(x_{0},y_{0},z_{0})^{T}$ and $K=L=3$.

\begin{figure}[t]%
\centering
\includegraphics[width=0.9\textwidth]{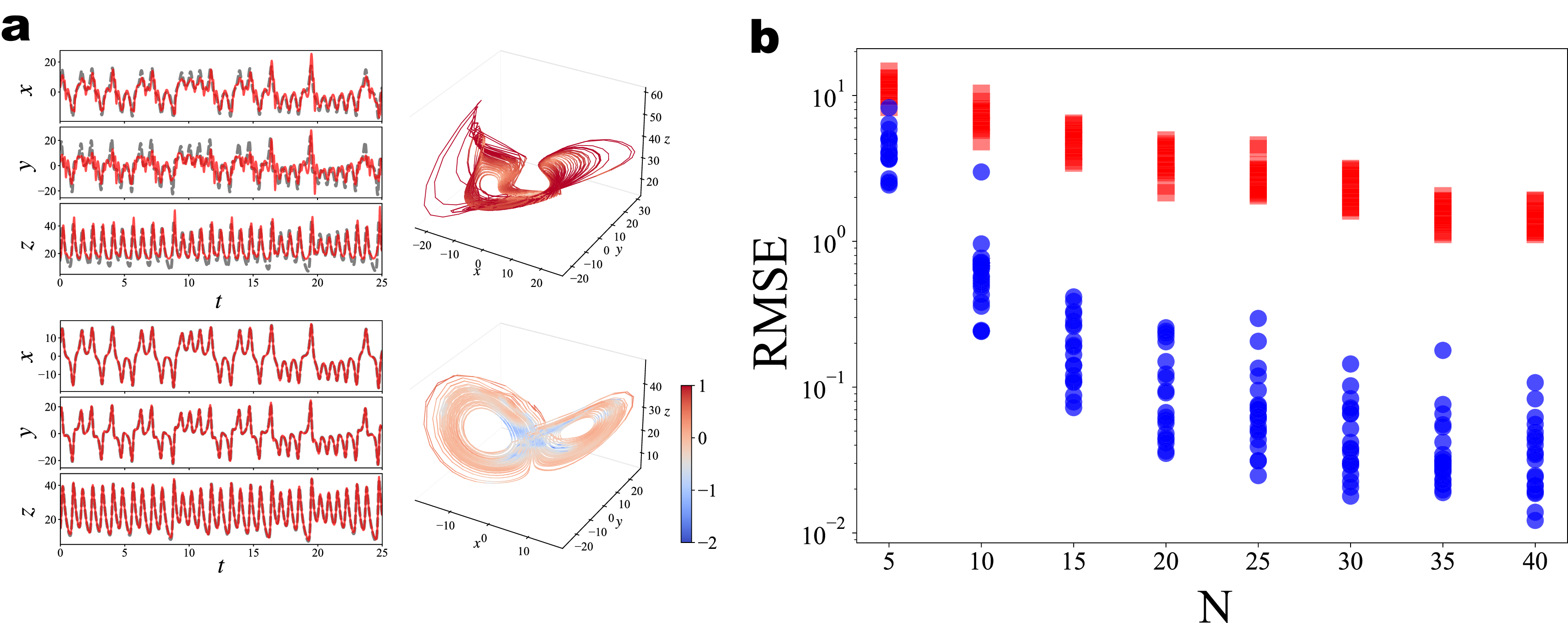}
\caption{
{\bf Short-term prediction (open-loop).}
{\bf a}, The top and bottom panels show the results using ${\mathcal{L}}$- and  ${\mathcal{Q}}$-ESN, respectively. The left and right panels show the time series of the target (grey dashed) and prediction (red solid)
and the phase space structures of the orbits.
The colors represent the local error of the prediction, $\| {\bf y}_{t} - \hat{\bf y}_{t} \|$.
{\bf b}, The root mean square error (RMSE),
$\sqrt{\langle \| {\bf y}_{t} - \hat{\bf y}_{t} \|^{2} \rangle_{T}}$, over the size of the network $N$.
The red and blue dots show the RMSE using ${\mathcal{L}}$- and  ${\mathcal{Q}}$-ESN, respectively. 
}\label{fig2}
\end{figure}

\subsection{Short-term prediction (open-loop)}
First, we study the short-term prediction, and in particular, $\tau=0.2$ ahead prediction of the Lorenz chaos; hence, when the input is ${\bf x}_{t}=\boldsymbol{\phi}^{t} ({\bf x}_{0})$,
the target output is ${\bf y}_{t}=\boldsymbol{\phi} ({\bf x}_{t})={\bf x}_{t+1}$.
%
Here, the random matrix $A$ is scaled so that its spectral radius is $0.95$, and $\sigma_{B}=0.1$ and $\beta=10^{-4}$ are used.

Fig.~2 a shows the prediction results with the ESN size $N=10$.
Note that while we also use $N=10$ in the long-term prediction later, which is quite small compared to the commonly used one, as a reference, $N=300$ is used in~\cite{pathak2017using}.
The left panels of Fig.~2 a are the time series of the target signals, i.e.,
${\bf y}_{t}$, depicted by the grey dashed lines,
and those of the predictions,
i.e.,
$\hat{\bf y}_{t}$, depicted by the red solid lines.
In the following, we refer to the conventional ESN as linear ESN (${\mathcal{L}}$-ESN) and the quadratic-form ESN as ${\mathcal{Q}}$-ESN.
The predictions by the ${\mathcal{L}}$-ESN and the ${\mathcal{Q}}$-ESN are shown in the upper and lower panels of Fig.~2 a, respectively.

Although there is a discrepancy between the target signal $\{ {\bf y}(t) \}$ and the prediction by the ${\mathcal{L}}$-ESN $\{ \hat{\bf y}(t) \}$, it is difficult to distinguish between the target signal $\{ {\bf y}(t) \}$ and the prediction by the ${\mathcal{Q}}$-ESN $\{ \hat{\bf y}(t) \}$, i.e., the ${\mathcal{Q}}$-ESN provides a more accurate prediction than the ${\mathcal{L}}$-ESN.
The right panels of Fig.~2 a show the phase space structures of the orbit $\{ \hat{\bf y}(t) \}$, corresponding to the time series on the left panels.
The phase space structures of the orbit $\{ \hat{\bf y}(t) \}$ by the ${\mathcal{L}}$-ESN are far from those of the true Lorenz attractor; however, the ${\mathcal{Q}}$-ESN can predict the orbit whose phase space structure is qualitatively the same as the true one, the butterfly wing shape.
In summary, the ${\mathcal{Q}}$-ESN is more accurate in short-term predictive ability than the ${\mathcal{L}}$-ESN when $N = 10$.

To quantitatively compare the predictive ability of the ${\mathcal{L}}$-ESN and the ${\mathcal{Q}}$-ESN, we plot the root mean square errors (RMSE),
$\sqrt{\langle \| {\bf y}_{t} - \hat{\bf y}_{t} \|^{2} \rangle_{T}}$, in Fig.~2 b for the ESN size $5 \le N \le 40$.
The values of the RMSE over the 20 different realizations for each case are shown to investigate the dependence of the random number realizations used for the matrices $A$ and $B$.
Here, the red and blue circles represent the RMSE given by the ${\mathcal{L}}$-ESN and the ${\mathcal{Q}}$-ESN, respectively.
While the RMSE values typically decrease with increasing ESN size, there is a huge gap between the RMSE values of the ${\mathcal{L}}$-ESN and the ${\mathcal{Q}}$-ESN.
In particular, for the case of $N=40$, the RMSE values of the ${\mathcal{Q}}$-ESN are $10^{-2}$ times those of the ${\mathcal{L}}$-ESN, i.e., the RMSE of the ${\mathcal{Q}}$-ESN is significantly lower than that of the ${\mathcal{L}}$-ESN.

\subsection{Long-term prediction (closed-loop)}
For the long-term prediction, the both ESNs are trained for the $\tau$-ahead prediction task where $\tau=0.02$ of the Lorenz chaos.
Again, the target output is ${\bf y}_{t} =\boldsymbol{\phi}( {\bf x}_{t})={\bf x}_{t+1}$.
The output from the ESN is denoted by $\hat{\bf y}_{t} = \hat{\bf h}( {\bf r}_{t})$,
where the output function is $\hat{\bf h}( {\bf r}_{t}) = W {\bf r}_{t}$ for the ${\mathcal{L}}$-ESN
and $\hat{\bf h}( {\bf r}_{t}) = W {\bf r}_{t} + {\bf r}_t^{T} W^{\mathcal{Q}} {\bf r}_{t}$ for the ${\mathcal{Q}}$-ESN.
After training, we obtain $\hat{\bf y}_{t} \simeq {\bf y}_{t} ={\bf x}_{t+1}$.
In the next step, we employ ${\bf r}_{t+1} = {\bf F}({\bf r}_{t}, \hat{\bf y}_{t}) = 
{\bf F}({\bf r}_{t}, \hat{\bf h}( {\bf r}_{t}) )=: {\bf G}({\bf r}_{t} )$
instead of (\ref{abstrc}),
where the map ${\bf G}: \mathbb{R}^{N} \to \mathbb{R}^{N}$ determines the {\it autonomous} dynamical system in the reservoir state space.
This closed-loop method using the ESN, which we call the {\it automated} ESN for short, not only provides long-term prediction, but also has the surprising ability to reconstruct the target dynamics determined by $\boldsymbol{\phi}$, as mentioned in the introduction.

Here we fix $N = 10$ and examine the effect of varying the functional form $\hat{\bf h}( {\bf r}_{t})$, i.e. using the ${\mathcal{L}}$-ESN or the ${\mathcal{Q}}$-ESN, on these abilities.
The random matrix $A$ is scaled so that its spectral radius is $0.01$, and $\sigma_{B}=0.01$ and $\beta=10^{-6}$ are used.

\begin{figure}[t]%
\centering
\includegraphics[width=0.9\textwidth]{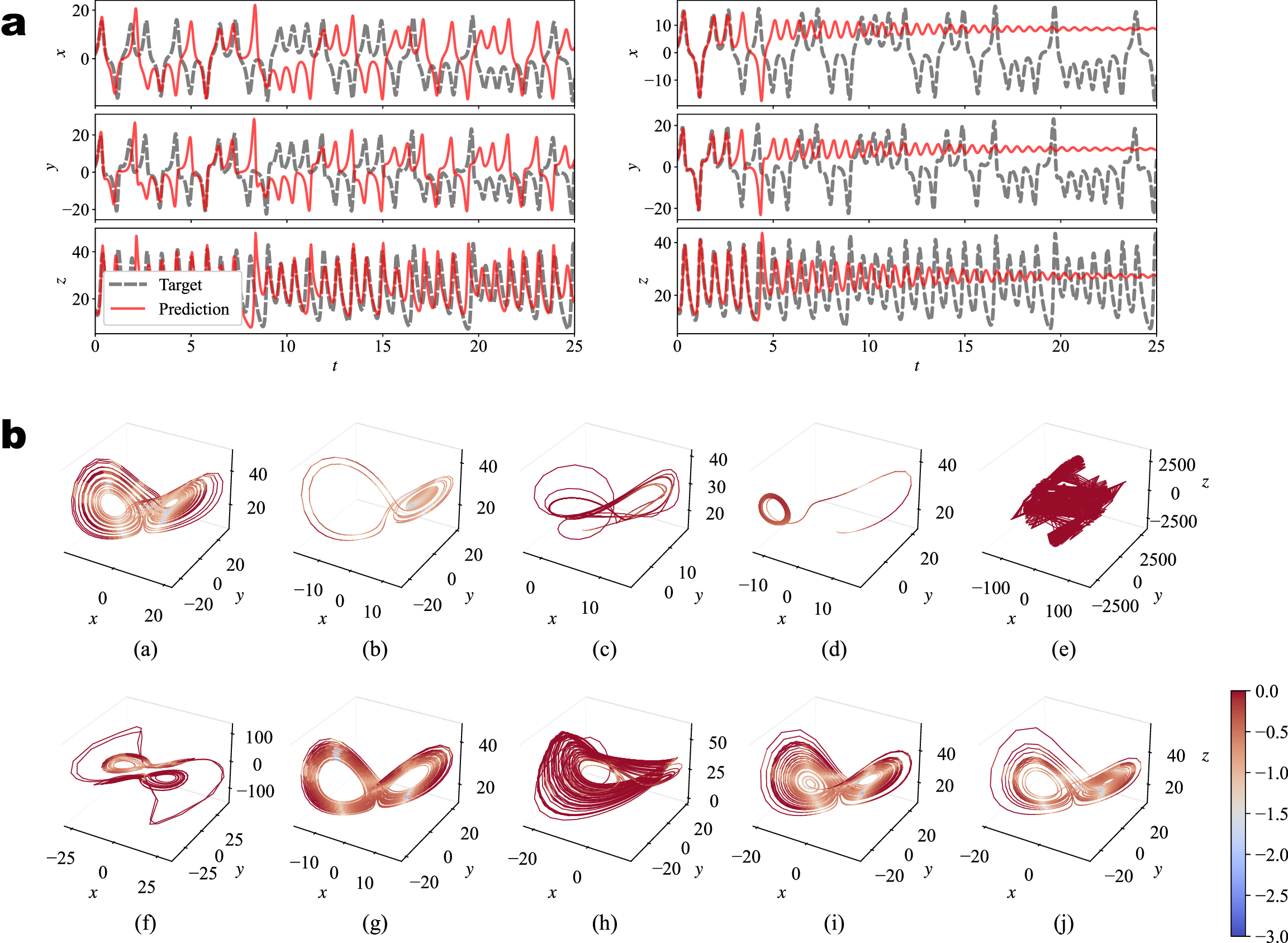}
\caption{
{\bf Long-term prediction (closed-loop) using the automated ${\mathcal{L}}$-ESN.}
{\bf a}, The time series of the target (grey dashed) and prediction (red solid).
The difference between the left and right panels lies in the realizations of the random numbers used for $A$ and $B$.
{\bf b}, The phase space structures of the orbits generated by the automated ${\mathcal{L}}$-ESN. The panels (a) -- (j) are results corresponding to the ten times realizations of the random numbers used for $A$ and $B$.
The colors represent the local conjugacy error, ${\mathcal{E}}^{c}_{t}$.
The length of the orbits shown is $T=50$.
}\label{fig3}
\end{figure}

\begin{figure}[t]%
\centering
\includegraphics[width=0.9\textwidth]{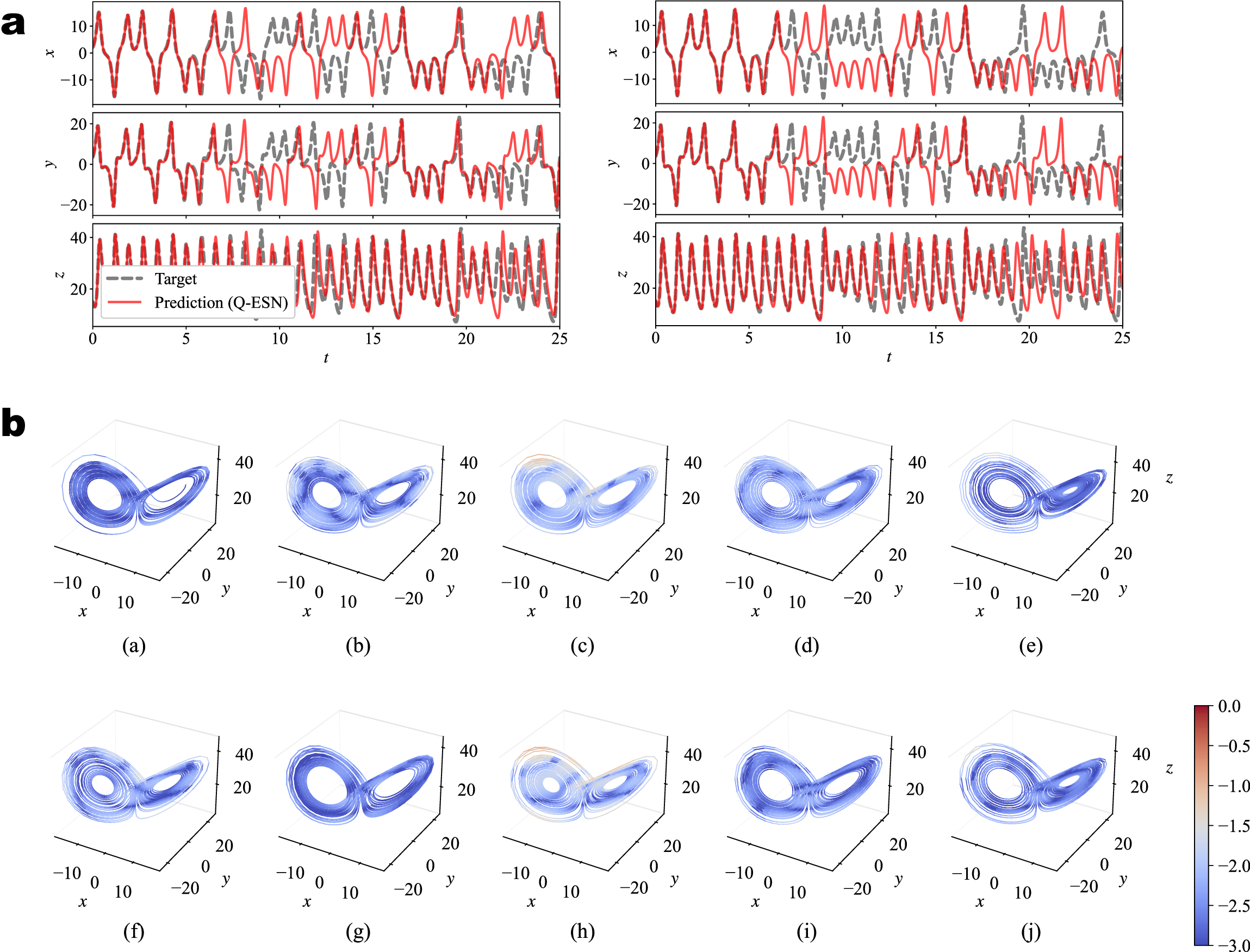}
\caption{
{\bf Long-term prediction (closed-loop) using the automated ${\mathcal{Q}}$-ESN.}
The same as Fig.~3, but the ${\mathcal{Q}}$-ESN is used instead of the ${\mathcal{L}}$-ESN.
}\label{fig1}
\end{figure}


First, Fig.~3 shows the results of long-term prediction using the automated ${\mathcal{L}}$-ESN.
The results depend on the realization of the random matrices $A$ and $B$,
we show two cases of different random numbers in Fig.~3 a,
where the black dashed lines and the red solid lines represent the time series of the target Lorenz chaos and the prediction by the automated ${\mathcal{L}}$-ESN, respectively.
For $t<0$ we use the open-loop method, and switch to the closed-loop method at $t=0$.
For both results, the orbits by the automated ${\mathcal{L}}$-ESN deviate significantly from the target orbits at $t \gtrsim 3$,
which is about three Lyapunov times, since the maximal Lyapunov exponent is $\lambda=0.906$.
The second case, shown in the right panel of Fig.~3 a, suggests that the dynamics generated by
the automated ${\mathcal{L}}$-ESN is not chaotic, but converges to a fixed point, which is unstable in the target Lorenz system.

To investigate the reconstruction ability, we demonstrate the phase space structure of the orbit generated by the automated ${\mathcal{L}}$-ESN for 10 different realizations in (a) -- (j) of Fig.~3 b,
where the left and right panels of Fig.~3 a correspond to the cases of (a) and (b), respectively.
Obviously, the reconstruction ability of the automated ${\mathcal{L}}$-ESN is highly dependent on the realizations; in other words, it is not robust.

Fig.~4 is the same as Fig.~3, but we use the ${\mathcal{Q}}$-ESN instead of the ${\mathcal{L}}$-ESN.
The automated ${\mathcal{Q}}$-ESN exhibits the long-term prediction ability over about 8 Lyapunov times, which is remarkably improved compared to the case of the ${\mathcal{L}}$-ESN used, even with the same network size.
Due to the intrinsic orbital instability of the Lorenz chaos, the orbits generated by the automated ${\mathcal{Q}}$-ESN inevitably deviate from the target orbits; however, the phase space structures shown in Fig.~4 b are qualitatively equivalent to the Lorenz attractor.
As will be quantitatively examined later, the reconstruction ability of the automated ${\mathcal{Q}}$-ESN is independent of the realizations; in other words, it can {\it robustly} reproduce the dynamics of Lorenz chaos.
Similar results have been reported in the previous studies, e.g. Pathak {\it et al.} (2017)~\cite{pathak2017using};
however, they used the relatively large network such as $N = 300$.
We emphasize that the automated ${\mathcal{Q}}$-ESN has such a long-term prediction and robust reconstruction ability with the tiny network, $N=10$.

\subsubsection{Quantitative comparison}

For quantitative comparison, we introduce the mean conjugacy error (henceforth MCE),
and the Kullback-Leibler divergence (henceforth KLD), which quantify the error between orbits and the error between invariant distributions, respectively.
First, we define the MCE.
As discussed in Hara and Kokubu (2022)~\cite{Kokubu2022learning}, the dynamical system determined by the automated ESN, $({\bf G}, \mathbb{R}^{N})$, is expected to be smoothly conjugate to the Lorenz dynamics, (${\boldsymbol{\phi}}, \mathbb{R}^{K}$).
The MCE quantifies the deviation from the expected conjugacy as follows.
The above relationship $\hat{\bf y}_{t} = \hat{\bf h} ({\bf r}_{t}) \simeq {\bf y}_{t} ={\bf x}_{t+1} = \boldsymbol{\phi}({\bf x}_{t})$ implies
\begin{align}
\hat{\bf h} ({\bf r}_{t}) \simeq \boldsymbol{\phi}({\bf x}_{t}) \simeq \boldsymbol{\phi} (\hat{\bf h} ({\bf r}_{t-1})).
\end{align}
On the other hand, for the automated ESN, we have ${\bf r}_{t}={\bf G}({\bf r}_{t-1})$, leading to
\begin{align}
\hat{\bf h} ({\bf r}_{t}) = \hat{\bf h} ({\bf G}({\bf r}_{t-1})).
\end{align}
Considering the map $\hat{\bf h}$ as the conjugacy map, we define the conjugacy error
at the reservoir state ${\bf r}_{t} \in \mathbb{R}^{N}$ by
\begin{align}
{\mathcal{E}}^{c}_{t} := \| \boldsymbol{\phi} (\hat{\bf h} ({\bf r}_{t})) - \hat{\bf h} ({\bf G}({\bf r}_{t})) \|,
\end{align}
where $\boldsymbol{\phi}$ is approximated by the four-stage and fourth-order Runge-Kutta method.
The long-time average of ${\mathcal{E}}^{c}_{t}$ along the orbit $\{ {\bf r}_{t} \}$ generated by ${\bf G}$ defines
the MCE,
\begin{align}
\bar{\mathcal{E}}^{c} := \langle {\mathcal{E}}^{c}_{t}  \rangle_{T} \label{def_MCE}
\end{align}

The colors of the orbits in Fig.~3 b and Fig.~4 b represent the conjugacy error, where the values on the color bars correspond to $\log {\mathcal{E}}^{c}_{t}$.
Obviously, compared to the automated ${\mathcal{L}}$-ESNs (Fig.~3 b), the colors of the orbits generated by the automated ${\mathcal{Q}}$-ESNs (Fig.~4 b) are blue almost everywhere on the attractor, suggesting that successful conjugacy to the target Lorenz dynamics.

While the MCE quantifies the reconstruction ability of the automated ESN, it is not perfect.
For instance, if the orbit generated by ${\bf G}$ converges to the saddle point along the stable manifold of the target system, the MCE may take a small value. However, the saddle point cannot be an attractor. Therefore, in this case, the automated ESN fails to reproduce the target attractor, even if the MCE is small.
To shed light on the ergodic aspect of the dynamics, we compare the invariant probability measures through $D_{\text{KL}}$ as a quantification complementary to MCE.

\begin{figure}[t]%
\centering
\includegraphics[width=0.9\textwidth]{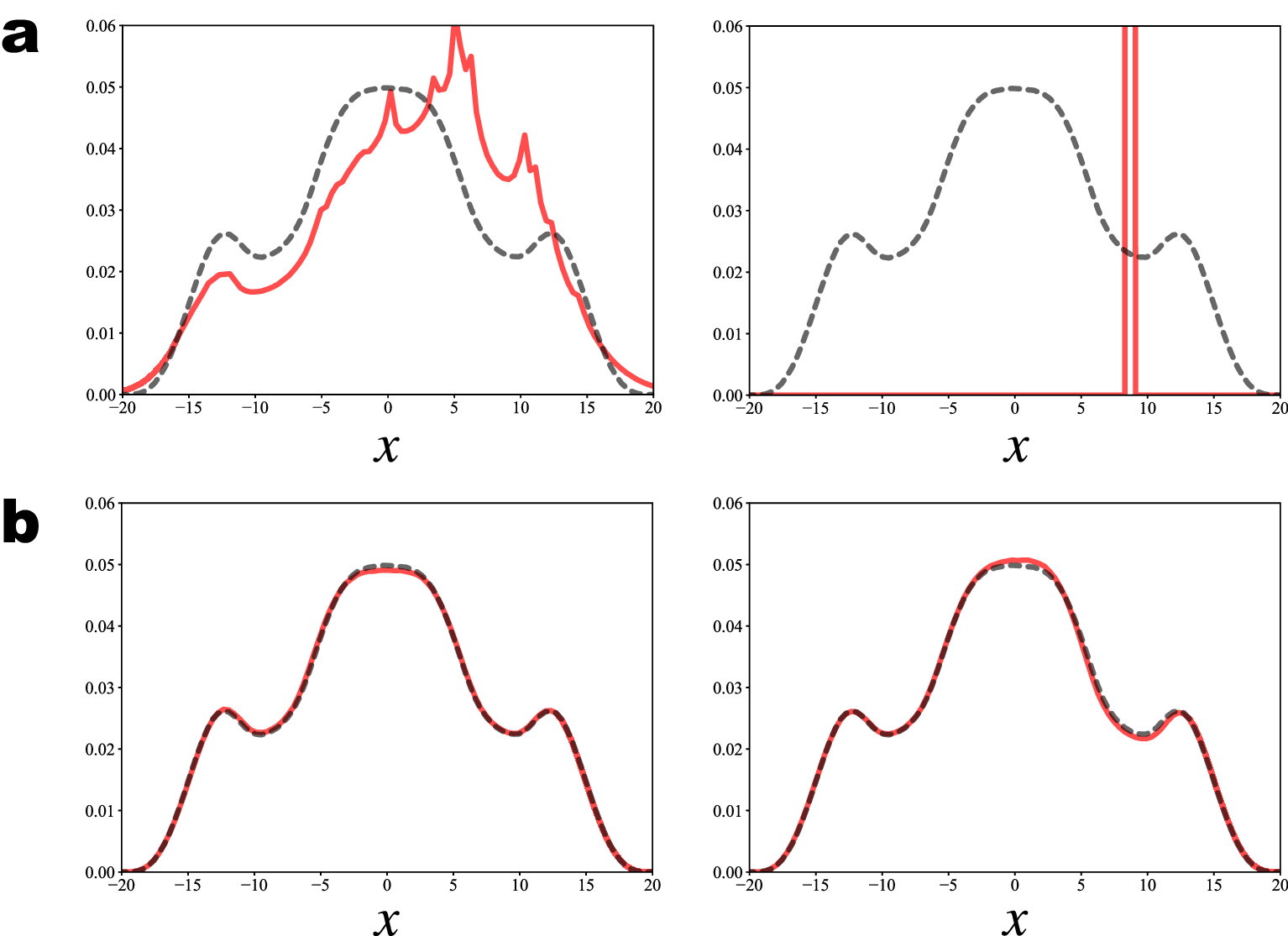}
\caption{
{\bf Probability density functions (PDF) of the variable $x$.} The dashed lines show the PDF of the target Lorenz system $p(x)$.
The red solid lines show the PDF $q(x)$ calculated from data generated by {\bf a}, ${\mathcal{L}}$-ESN and {\bf b}, ${\mathcal{Q}}$-ESN. Two panels of {\bf a} and {\bf b} correspond to the cases shown in Fig.~3 a and Fig.~4 a, respectively.
}\label{fig1}
\end{figure}

Fig.~5 shows the probability density functions (PDF) of the variable $x$.
The grey dashed and red solid lines represent the PDF $p(x)$ calculated from the target Lorenz chaos data and the PDF $q(x)$ calculated from the automated ESN data, respectively.
The two panels of Fig.~5 a show the results of the automated ${\mathcal{L}}$-ESN, corresponding to the two cases shown in Fig.~3 a.
Although the time series shown in the left panel of Fig.~3 a
and the phase space structure shown in Fig.~3 b (a) are similar to the target Lorenz,
the PDF $q(x)$ shown in the left panel of Fig.~5 a differs from $p(x)$.
The time series shown in the right panel of Fig.~3 a converges to the fixed point, resulting in the PDF $q(x)$ shown in the right panel of Fig.~5 a having a delta function-like form, and apparently differing from $p(x)$.

The two panels of Fig.~5 b show the results of the automated ${\mathcal{Q}}$-ESN, corresponding to the two cases shown in Fig.~4 a.
The PDFs $p(x)$ and $q(x)$ are quite similar, suggesting that the automated ${\mathcal{Q}}$-ESN can reproduce the global structure of the target Lorenz attractor, in addition to the accurate prediction along the orbit verified by the MCE, which is local in phase space.
The values of $D_{\text{KL}}$, where its definition is $D_{\text{KL}}(p \| q):= - \int p(x) \ln (q(x)/p(x)) dx$, are $D_{\text {KL}}(p\| q)=6.4 \times 10^{-5}$ and $1.6 \times 10^{-4}$ for the PDFs $q(x)$ shown in the left and right panels of Fig.~5 b, respectively.
These values of $D_{\text{KL}}$ are significantly smaller than the values of $D_{\text{KL}}$ in the case of the ${\mathcal{L}}$-ESN used, e.g. 
$D_{\text{KL}}(p\| q)=4.8 \times 10^{-2}$ for the PDFs $q(x)$ shown in the left panel of Fig.~5 a.

\begin{figure}[t]%
\centering
\includegraphics[width=0.9\textwidth]{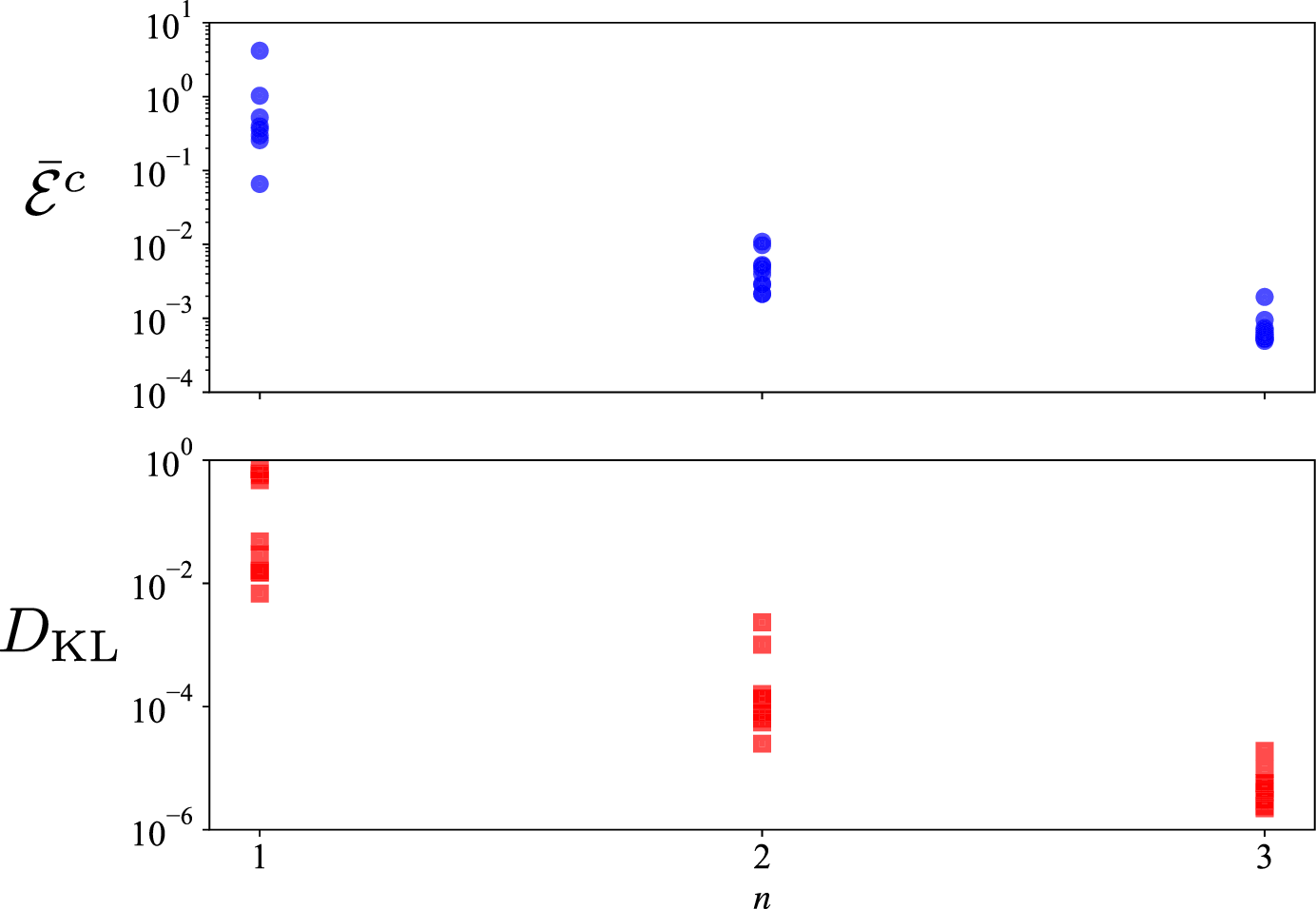}
\caption{
{\bf Summary of quantitative comparison of reconstruction ability}.
The left ($n=1$), center ($n=2$), and right ($n=3$) points correspond to the results of
${\mathcal{L}}$-,
${\mathcal{Q}}$-, and
${\mathcal{C}}$-ESN, respectively.
The top and bottom panels show the MCE $\bar{\mathcal{E}}^c$ and KLD $D_{\text{KL}}$ values over ten times realizations of the random matrices $A$ and $B$, respectively.}
\label{Fig6}
\end{figure}

Fig.~6 summarizes the quantitative comparison. 
The top and bottom panels show the MCE $\bar{\mathcal{E}}^c$ and KLD $D_{\text{KL}}$ values over ten times realizations of the random matrices $A$ and $B$, respectively.
The left points, labelled ``$1$'' on the horizontal axis,
represent the results for the automated ${\mathcal{L}}$-ESN, excluding the two extremely poor results shown in Fig.~3 b (e) and (f).
The centre points, labelled ``$2$'' on the horizontal axis,
 represent the results for the automated ${\mathcal{Q}}$-ESN, illustrating that the values of both the MCE $\bar{\mathcal{E}}^c$ and the KLD $D_{\text{KL}}$ are significantly smaller than those of the ${\mathcal{L}}$-ESN.
Immediately we notice this remarkable reconstruction ability of the ${\mathcal{Q}}$-ESN.
Moreover, we find that both quantities are less dependent on the realizations of the random matrices compared to the ${\mathcal{L}}$-ESN, i.e. the ${\mathcal{Q}}$-ESN improves not only the accuracy but also the robustness of the reconstruction results.

Finally, we remark the results beyond the ${\mathcal{Q}}$-ESN, i.e. the cubic-form ESN (${\mathcal{C}}$-ESN) including up to the third order terms of the reservoir variables, ${\bf r}_{j} {\bf r}_{k} {\bf r}_{\ell}$, and the corresponding output weights, $W_{ijk\ell}^{\mathcal{C}}$, which are trained to approximate the third term in the Taylor expansions (\ref{Taylor}) as
\begin{align}
W_{ijk\ell}^{\mathcal{C}} \simeq \frac{1}{6} \frac{\partial^{3} {\bf h}_{i}({\bf 0})}{\partial {\bf r}_{j} \partial {\bf r}_{k} \partial {\bf r}_{\ell}}.
\end{align}
We show the MCE $\bar{\mathcal{E}}^c$ and the KLD $D_{\text{KL}}$ for the automated ${\mathcal{C}}$-ESN in the right points, labelled ``$3$'' on the horizontal axis, of Fig.~6.
As expected, the accuracy of the reconstruction by the automated ${\mathcal{C}}$-ESN is superior to that of
the ${\mathcal{L}}$-ESN and the ${\mathcal{Q}}$-ESN.
Furthermore, we find again that the ${\mathcal{C}}$-ESN improves not only the accuracy but also the robustness of the reconstruction results, compared to those of the ${\mathcal{L}}$-ESN and the ${\mathcal{Q}}$-ESN.

\section{Conclusion and Discussion}

Inspired by the seminal works on the mathematical analysis of RC~\cite{grigoryeva2021chaos,Kokubu2022learning}, we have proposed a novel method of RC with generalized readout with a theoretical guarantee of its high computational capabilities based on generalized synchronization.
Numerical studies on the Lorenz chaos have uncovered significant short- (Fig.~2) and long-term prediction and reconstruction abilities with improved robustness (Fig.~3 -- 5) of the ${\mathcal{Q}}$-ESN.
The MCE $\bar{\mathcal{E}}^c$ and KLD $D_{\text{KL}}$ have quantified these properties complementarily, i.e. from the notions of orbit and distribution.
By including the higher-order approximation, we have revealed ``hierarchical'' improvement in reconstruction ability and robustness; i.e. the ${\mathcal{C}}$-ESN is superior to the ${\mathcal{Q}}$-ESN, which is superior to the ${\mathcal{L}}$-ESN (Fig.~6).

As the future extensions based on the present work, we discuss the following three directions: mathematical analysis, machine learning, and physical implementation.
From the mathematical analysis of RC~\cite{grigoryeva2021chaos,Kokubu2022learning}, it may be natural that
introducing the generalized readout improves prediction ability.
However, we unexpectedly observed an improvement in the robustness of the reconstruction ability.
Further analysis of the reservoir dynamics is crucial; unveiling fundamental properties such as the topological conjugacy and the mechanism behind the enhanced robustness will have major implications for several fields, including machine learning, where stabilizing the dynamics of neural networks by adding noise and normalization is one of the critical issues~\cite{wikner2024stabilizing}.

The generalized readout paves the way for novel physical implementations of RC. Among the various implementations, some physical RC can utilize only a highly constrained degree of freedom~\cite{sunada2019photonic,appeltant2011information,wang2023echo,takano2018compact,tanaka2019recent},
e.g. increasing the number of virtual nodes is essential for RC with a photonic integrated circuit~\cite{takano2018compact}.
Even if only a highly constrained degree of freedom is available, the generalized readout can construct a rich basis for learning from such low-dimensional dynamics, as shown in the numerical study where $N=10$.

The apparent drawback of using the generalized readout is the large number of parameters to be trained, still within the linear learning framework. Therefore, based on the hierarchical improvement in accuracy with increasing parameters (Fig.~6), the balance between accuracy and learning cost should be determined for each application.
For the large number of parameters to be trained, transfer learning~\cite{inubushi2020transfer,sakamaki2022transfer} may be efficient.
Once linear regression is used, the trained parameters, i.e. the generalized readout weights, can be reused with a minor correction for similar tasks, e.g. predicting chaotic dynamics that are structurally stable.

The automated RC with generalized readout achieves accurate predictions, e.g. longer than 8 Lyapunov times (Fig.~4); however, such a prediction eventually fails due to the orbital instability.
Toward practical predictions of, for instance, fluid turbulence~\cite{inubushi2023characterizing}, the integration of the automated RC with generalized readout and data assimilation may be essential in future work.
The concept of generalized readout does not require the RC framework, but rather, may be essential in a more general machine learning context, e.g. training recurrent neural networks. 
Studies of neural connections similar to ${\mathcal{Q}}$- and ${\mathcal{C}}$-ESN may also be interesting in the context of a learning mechanism in biological brains.
Whatever the direction, the concepts from dynamical system theory used in the above discussion, such as synchronization, orbital instability, and conjugacy, will shed light on a guiding principle for future studies.

\bmhead{Acknowledgments}
We thank M.~Hara, H.~Kokubu, S.~Sunada, T.~Yoneda, S.~Matsumoto, S.~Goto, Y.~Saiki, and J.~A.~Yorke for their insightful comments and encouragement. We would also like to thank C.~P.~Caulfield and DAMTP, University of Cambridge, for providing a great environment in which this work was completed on M.I.'s sabbatical.
This work was partially supported by JSPS Grants-in-Aid for Scientific Research (Grants No.~22K03420, No.~22H05198, No.~20H02068, and No.~19KK0067).




\bibliography{bibGRC}
\bibliographystyle{bst/sn-nature}

\end{document}